# Improving Robustness of Deep Convolutional Neural Networks via Multiresolution Learning

Hongyan Zhou, Yao Liang

*Abstract*—**The current learning process of deep learning, regardless of any deep neural network (DNN) architecture and/or learning algorithm used, is essentially a single resolution training. We explore multiresolution learning and show that multiresolution learning can significantly improve robustness of DNN models for both 1D signal and 2D signal (image) prediction problems. We demonstrate this improvement in terms of both noise and adversarial robustness as well as with small training dataset size. Our results also suggest that it may not be necessary to trade standard accuracy for robustness with multiresolution learning, which is, interestingly, contrary to the observation obtained from the traditional single resolution learning setting.**

*Index Terms*—**Multiresolution deep learning, Robustness, Wavelet, 1D/2D signal classification**

## I. INTRODUCTION

Deep neural network (DNN) models have achieved breakthrough performance on a number of challenging problems in machine learning, being widely adopted for increasingly complex and large-scale problems including image classification, speech recognition, language translation, drug discovery, and self-driving vehicles (e.g., [1], [2], [3], [4]). However, it has been revealed that these deep learning models suffer from brittleness, being highly vulnerable to even small and imperceptible perturbations of the input data known as adversarial examples (see the survey [5] and the references therein). Remarkably, a recent study has shown that perturbing even one pixel of an input image could completely fool DNN models [6]. Robustness in machine learning is of paramount importance in both theoretical and practical perspectives.

This intrinsic vulnerability of DNN presents one of the most critical challenges in DNN modeling, which has gained significant attention. Existing efforts on improving DNN robustness can be classified into three categories: (1) Brute-force adversarial training, i.e., adding newly discovered or generated adversarial inputs into training data set (e.g., [7, 8]), which dramatically increases training data size; (2) modifying DNN architecture, by adding more layers/subnetworks or designing robust DNN architecture (e.g., [9, 10]); and (3) modifying training algorithm or introducing new regularizations (e.g., [11, 12]). Unlike those existing works, we take a different direction. We focus on how to improve the learning process of neural networks in general. The observation is that the current learning process, regardless of DNN architecture and/or learning algorithm (e.g., backpropagation) used, is essentially a single resolution training. It has been shown that for shallow neural networks (SNN) this traditional single resolution learning process is limited in its efficacy, whereas multiresolution learning can significantly improve models' robustness and generalization [13, 14]. We are interested in investigating if such multiresolution learning paradigm can also be effective for DNN models as well to address their found vulnerability. In particular, the original multiresolution learning was mainly focused on 1-demisional (1D) signal prediction, which needs to be extended for 2-demisioal (2D) signal prediction tasks such as image classification.

In this work, we show that the multiresolution learning paradigm is effective for deep convolutional neural networks (CNN) as well, particularly in improving the robustness of DNN models. We further extend the original multiresolution learning to 2D signal (i.e., image) prediction problem and demonstrate its efficacy. Our exploration also shows that with multiresolution learning it may not be necessary to have the so-called tradeoff

H. Zhou is with the Department of Computer and Information Science, Indiana University Purdue University, Indianapolis, IN 46202 USA (e-mail: zhou94@iu.edu).

Y. Liang is with the Department of Computer and Information Science, Indiana University Purdue University, Indianapolis, IN 46202 USA (e-mail: yaoliang@iu.edu). Corresponding author.

between the accuracy and robustness, which seems contrary to the recent findings of [15] in traditional single resolution learning setting.

## II. Multiresolution Learning

### A. The main idea

The multiresolution learning paradigm for neural network models was originally proposed in [13, 14], and was demonstrated in applications to VBR video traffic prediction [16, 17, 18]. It is based on the observation that the traditional learning process, regardless of the neural network architectures and learning algorithms used, is basically a *single-resolution learning* paradigm. From multiresolution analysis framework [19] in wavelet theory, any signal can have a multiresolution decomposition in a tree-like hierarchical way. This systematic wavelet representation of signals directly uncovers underlying structures and patterns of individual signals at their different resolution levels, which could be hidden or much complicated at the finest resolution level of original signals. Namely, any complex signal can be decomposed into a simplified approximation (i.e. a coarser resolution) signal plus a "detail" signal which is the difference between the original and the approximation signals. A low pass filter $L$ is used to generate the coarse approximation signal (i.e., Low frequency component) while a high pass filter $H$ is used to generate the detail signal (i.e., High frequency component) as shown in Equations (1) and (2), respectively, where $f^i$ denotes the signal approximation at resolution level $i$ ($i \in Z$ and $i \geq 0$), and $d^i$ denotes the detail signal at resolution level $i$. This decomposition process can be iterated on approximation signals until an appropriate level of approximation is achieved. For the sake of notation convenience, $L^0$ (Layer 0) denotes the original signal $f^m$ at the finest resolution level $m$, whereas $L^i$ and $H^i$ denote approximation $f^{m-i}$ and detail $d^{m-i}$ signals, respectively, at Layer $i$ of signal decomposition.

$$f^{i-1} = Lf^i, \ i = 1, 2, ..., m. \tag{1}$$

$$d^{i-1} = Hf^i, \ i = 1, 2, ..., m. \tag{2}$$

This signal decomposition is reversable, meaning that a coarser resolution approximation plus its corresponding detail signal can reconstruct a finer resolution version of the signal recursively without information loss. One level of signal reconstruction is as follows:

$$f^i = f^{i-1} \langle + \rangle d^{i-1}, i = 1, 2, ..., m. \tag{3}$$

where operator $\langle + \rangle$ denotes the reconstruction operation. Fig. 1 shows the decomposition hierarchy of signal $f^i$ from the original finest resolution level $m$ to resolution level $m$-4. As we can see, by applying low pass filter $L$ and high pass filter $H$ on signal $L^0(f^m)$, we obtain $L^1$ and $H^1$. Recursively, $L^1$ is further deconstructed into $L^2$ and $H^2$, and so on.

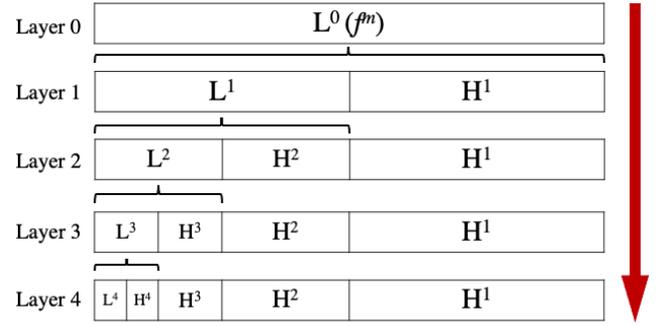

Fig. 1. Illustration of decomposition structure of 1D signal $f^m$ from the original finest resolution to lower resolution levels.

Utilizing the mathematical framework of multiresolution analysis [19], multiresolution learning [13, 14] explores and exploits the different signal resolutions in neural network learning, where the entire training process of neural network models is divided into multiple learning phases associated with different resolution approximation signals, from the coarsest version to finest version. This is in contrast with the traditional learning process widely employed in today's deep learning, where only the original finest resolution signal is used in the entire training process. The main idea of multiresolution learning is that for models to learn any complex signal better, it would be more effective to start training with a significantly simplified approximation version of the signal with a lot of details removed, then progressively adding more and more details and thus refining the neural network models as the training proceeds. This paradigm embodies the principle of divide and conquer applied to information science.

In multiresolution learning, discrete wavelet transform (DWT) can be employed for signal decomposition, whereas inverse DWT (IDWT) can be used to form simplified approximation signals via the replacement of the detail signal by zero signal in the signal reconstruction. To build a neural network model for a given task of sampled signal $f^m$, multiresolution learning can be outlined as follows [13, 14].

First, decompose the original signal $f^m$ to obtain $f^{m-1}, f^{m-2}$, ..., etc., based on which one reconstructs multiresolution versions of training data, where each different resolution approximation should maintain the dimension of the original signal that matches the input dimension of neural network model. Namely, reconstruct $k$ multiresolution versions $r_i$ ($i = 1, 2, ..., k$) of the original signal training data, where the representation of training data $r_i$ at resolution version $i$ is formed as follows:

$$r_i = \begin{cases} f^m, & i = 1 \\ (... ((f^{m-i+1} \langle + \rangle 0^{m-i+1}) \langle + \rangle 0^{m-i+2}) \langle + \rangle ... \langle + \rangle 0^{m-1}), & i > 1 \end{cases} \tag{4}$$

where $0^j$ indicates a zero signal at resolution level $j$ in multiresolution analysis.

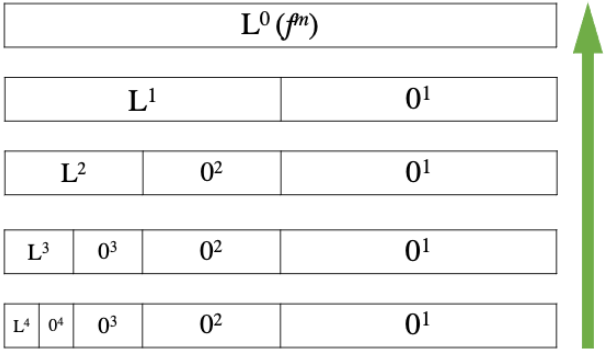

Fig. 2. Illustration of the information content of 1D signal training data at five different resolution levels, which can be organized into five information layers. The top layer (Layer 0) corresponds to original signal at the finest resolution level $m$, whereas the bottom layer (Layer 4) indicates the information content contained in the generated training data at the coarsest resolution level $m - 4$. The coarser resolution, the less detailed information it contains.

Fig. 2 illustrates how information content varies in $k$ multiresolution signal training data generated by (4) from the original (finest resolution) training data for $k = 5$.

Second, the training process of a neural network model is divided into a sequence of learning phases. Let $A_i(r_i)$ be a training phase conducted on training data version $r_i$ of a coarser resolution $i$ with any given learning algorithm. Let A→B indicate that A activity is carried out in advance of B activity. Thus, $k$-level multiresolution learning ($k > 1$) is formulated as an ordered sequence of $k$ learning phases associated with the sequence of approximation subspaces in multiresolution analysis, which satisfies $A_{i+1}(r_{i+1}) \rightarrow A_i(r_i)$ $(i=1, 2, ..., k-1)$ during the entire training process of a model.

It can be seen that the multiresolution learning paradigm forms an ordered sequence of $k$ learning phases stating with an appropriate courser resolution version of training data $r_k$, proceeding to the finer resolution versions of training data. The finest resolution of training data originally given will be used in the last learning phase $A_1(r_1)$. The total number of multiresolution levels $k$ is to be chosen by the user for a given task at hand. Also, the traditional (single resolution) learning process is just a special case of the multiresolution learning where the total multiresolution levels $k=1$.

### B. New Extension

While the original multiresolution learning [13, 14] was proposed for 1D signal prediction problems, it can be naturally extended for 2D signal prediction problems, such as image recognition. In this work, the original multiresolution learning paradigm is extended for 2D signal problems as follows. First, 2D multiresolution analysis is employed. Basically, 2D wavelet decomposition and its reconstruction are applied to construct coarser resolution approximation versions for 2D signal (training data). For example, 2D DWT can be performed iterating two orthogonal (horizontal and vertical) 1D DWT. For any 2D signal (image) at a given resolution level, four sub-band components are obtained by 2D DWT: the low-resolution approximation (sub-band $LL$) of the input image and the other three sub-bands containing the details in three orientations (horizontal $LH$, vertical $HL$, and diagonal $HH$) for this image decomposition.

Let $f^m$, a sampled 2D signal at a given resolution level $m$, be denoted as $LL^0$ (Layer 0) in 2D signal decomposition hierarchy. The coarser approximation $f^{m-1}$ obtained by 2D DWT, is then denoted as $LL^1$, with a one-quarter size of that of the original $f^m$; the other three sub-bands $d_{LH}^{m-1}$, $d_{HL}^{m-1}$, $d_{HH}^{m-1}$ are denoted as $LH^1$, $HL^1$ and $HH^1$, generated through two orthogonal 1D DWT in the order (L, H), (H, L) and (H, H), respectively. Similarly, this 2D signal decomposition process of $f^i$ ($i=m, m-1, ..., m-k+1$) can be iterated until an appropriate coarser approximation $LL^k$ (i.e., $f^{m-k}$) is achieved, as illustrated in Fig. 3. Again, this 2D signal decomposition is reversable. One level of 2D signal reconstruction is as follows:

$$f^i = f^{i-1}\langle+\rangle d_{LH}^{i-1}\langle+\rangle d_{HL}^{i-1}\langle+\rangle d_{HH}^{i-1}, \quad i = 1, 2, ..., m. \quad (5)$$

We can rewrite (5) as:

$$LL^j = LL^{j+1}\langle+\rangle LH^{j+1}\langle+\rangle HL^{j+1}\langle+\rangle HH^{j+1},$$
$$j = 0, 1, ..., m-1. \quad (6)$$

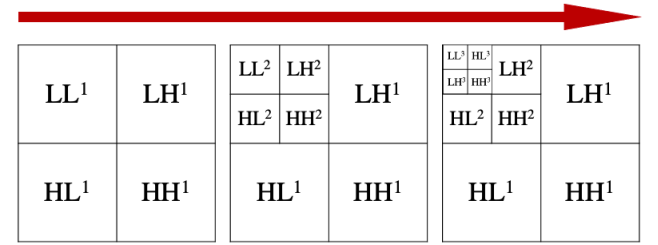

Fig. 3. Illustration of three-level wavelet decomposition of 2D signal $f^m$ ($LL^0$) from the given resolution level $m$.

To construct multiresolution versions of training data, we create one additional intermediate resolution level for each level of 2D decomposition to reduce information difference between resolutions, as shown in the second row in Fig. 4. This is one significant difference from the multiresolution learning procedure for 1D signal problem. Hence, the representation of multiresolution training data $r_i^{2D}$ at resolution level $i$ is constructed as follows:

$$r_i^{2D} =$$
$$\begin{cases} f^m, & i = 1 \\ (... ((LL^k\langle+\rangle 0^k)\langle+\rangle 0^{k-1}) ... \langle+\rangle 0^1), & i = 2k+1 \\ (... ((LL^k\langle+\rangle LH^k\langle+\rangle HL^k\langle+\rangle 0_{HH}^k)\langle+\rangle 0^{k-1}) ... \langle+\rangle 0^1), & i = 2 \end{cases}$$
$$(7)$$

where $0^j = 0_{LH}^j \langle+\rangle 0_{HL}^j \langle+\rangle 0_{HH}^j$, and integer $k > 0$.

Fig. 4 illustrates the information content of 2D signal training data at five different resolutions. The coarser resolution version of training data, the less detailed information it contains. Fig. 5 gives an example of four different resolution training data constructed from (7).

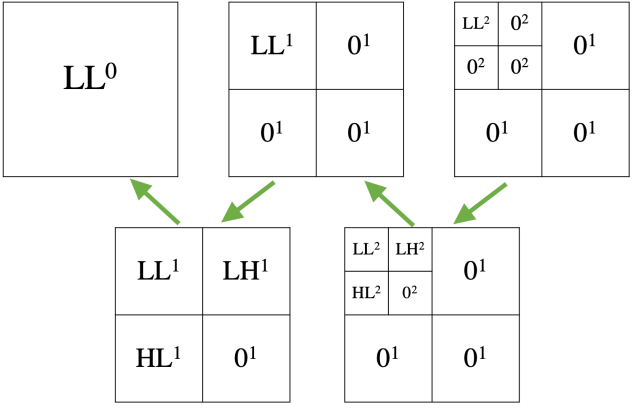

Fig. 4. Illustration of the information content of 2D signal training data at five different resolution levels, indicated by arrows from the coarsest resolution to the finest resolution. The coarser resolution, the less detailed information it contains.

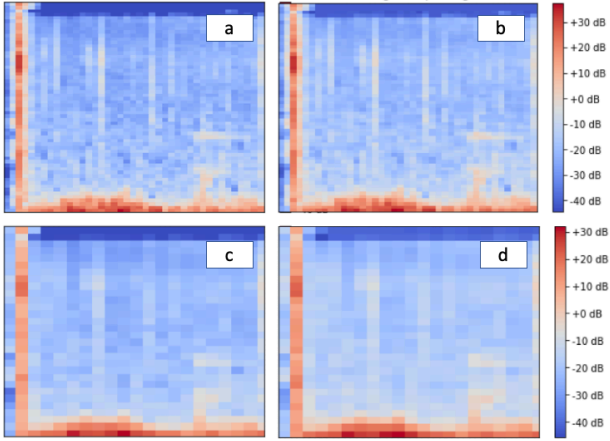

Fig. 5. An example of a sound sample of log-mel spectrogram represented in four resolution levels: (a) $r_1^{2D}$, (b) $r_2^{2D}$, (c) $r_3^{2D}$, and (d) $r_4^{2D}$.

### C. Multiresolution Learning Process

All the multiresolution training data versions are of the same dimension size but associated with different detailed levels of the given signal. For 1D signals, each level of wavelet decomposition will create one new version of coarser resolution training data, while for 2D signals, each level of wavelet decomposition will create two new versions of coarser resolution training data. As the version index $i$ of training data $r_i/r_i^{2D}$ increases, the $r_i/r_i^{2D}$ contain less details, which facilitates to start model training with a significantly simplified signal version even though the original signal could be extremely complex. Furthermore, one has the flexibility to select an appropriate wavelet transform and basis for signal decomposition and a total multiresolution levels $k$ in multiresolution learning for a given task. After each learning phase in multiresolution learning, the resulting intermediate weights of the DNN model are saved and used as initial weights of the model in the next learning phase in this dynamic training process, until the completion of the final learning phase with the original finest resolution training data $r_1/r_1^{2D}$ ($f^m$).

### III. EXPERIMENTS AND ANALYSES

We conduct thorough experiments with three open datasets FSDD [20], ESC-10 [21] and CIFAR-10 [22] of audio/image to systematically study and compare the multiresolution learning with the traditional single resolution learning on deep CNN models. Particularly, the problem of speech recognition is casted, respectively, to 1D signal classification on data set FSDD, and 2D image classification on data set ESC-10, to illustrate our extended multiresolution learning. The following subsections describe the data sets used, our experiment setup, and results and evaluation.

### A. Datasets and Preprocessing

The FSDD [20] is a simple audio/speech dataset consisting of sample recordings of spoken digits in wav files at 8kHz. FSDD is an open dataset, which means it will grow over time as new data are contributed. In this work, the version used consists of six speakers, 3,000 recordings (50 of each digit per speaker), and English pronunciations. Even though the recordings are trimmed so that they have near minimal silence at the beginning and end, there are still a lot of zero values in the raw waveform of the recordings.

To apply DNN models to speech recognition, some preprocessing steps are necessary for raw speech samples, which include cropping, time scaling, and amplitude scaling. First, in the cropping step, all contiguous zero values at the beginning and the end of recordings should be removed, to reserve the significant signal part. However, that means the recordings can have different lengths after cropping. Second, in the time scaling step, a fixed-length $L$, a number near the mean value of lengths of all sample recordings, is chosen as the dimension of input layer of DNN models; each sample recording of variable length is then either extended or contracted to get the same length $L$. Third, in the amplitude scaling step, the amplitude of each recording is normalized to a range of [-1,1]. Fig. 6 shows six examples of three different preprocessing steps. The leftmost shows the original signals before processing. After cropping all the zero values in the beginning and end, signals are shown in the middle part. The rightmost gives the final preprocessing results after scaling time steps and amplitude. Scaling time helps obtain a unified signal size as model input size while scaling amplitude reduces the influence of different loudness levels since loudness is not our focus.

The ESC-10 dataset [21] contains sounds that are recorded in numerous outdoor and indoor environments. Each sound sample recording is of 5 second duration. The sampling rate of each recording is 44100 Hz. The dataset consists of a total of 400 recordings, which are divided into 10 major categories. They are sneezing, dog barking, clock ticking, crying baby, crowing rooster, rain, sea waves, fire crackling, helicopter, and chainsaw. Each category contains 40 samples. The dataset comes with 5 folds for cross-validation.

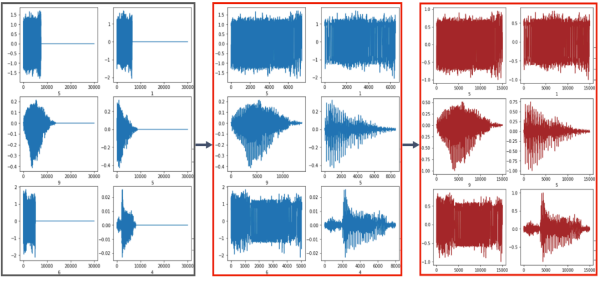

Fig. 6. Illustration of preprocessing steps. The left-most column shows the original signals. The middle column shows obtained signals after cropping all the zero values at the beginning and end of recordings. The rightmost column shows the results after both scaling time and scaling amplitude.

To apply multiresolution learning on the 2D CNN for speech recognition, log-scaled mel-spectrograms (i.e., images) were first extracted from all sound recordings with a window size of 1024, hop length of 512, and 60 mel-bands. Fig. 7 illustrates the generated spectrogram of two examples.

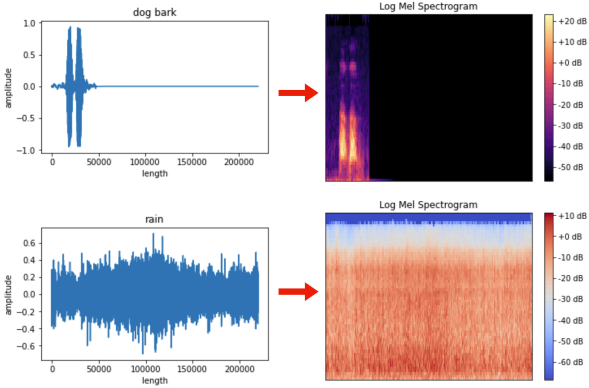

Fig. 7. Spectrograms extracted from a clip (dog bark) and a clip (rain).

Some environmental sound clips have periodic features. For example, we can hear bark several times or tick several times in a single sample recording. Also, considering the very limited number of samples available for training, the spectrograms were split into 50% overlapping segments of 41 frames (short variant, segments of approx. 950 ms) [21], but without discarding silent segments. One example is shown in Fig. 8. The top image shows the original spectrogram extracted from a clip (dog bark), while the second, third, and fourth images illustrate the first three short segments split from the original spectrogram, where the overlaps about 50% with neighboring short segments can be clear seen.

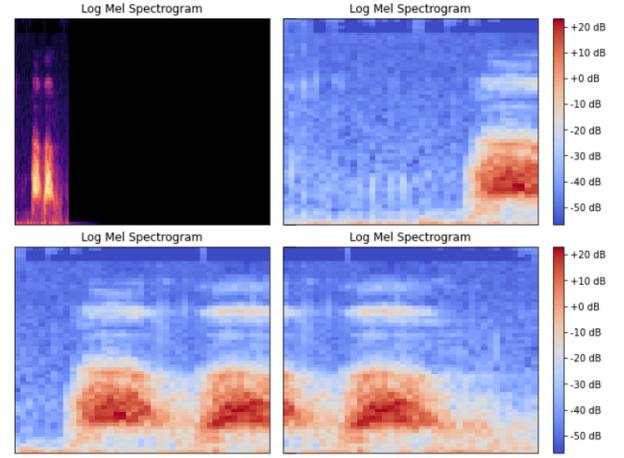

Fig. 8. Spectrogram clipping. The first image shows the original spectrogram extracted from a clip (dog bark). The second, third, and fourth images illustrate the first 3 short segments split from the original spectrogram.

The CIFAR-10 dataset [22] contains 50,000 training and 10,000 testing 32x32 color images of 10 classes. Both training set and testing set are normalized with mean and standard deviation respectively. Considering the small size of images in CIFAR-10, we apply multiresolution learning with three levels of only transform original training images into another two resolutions, which means resolutions $r_1^{2D}$, $r_2^{2D}$ and $r_3^{2D}$. will be fed into the training network. Fig. 820 shows several sample examples of the transformation results before applying normalization method.

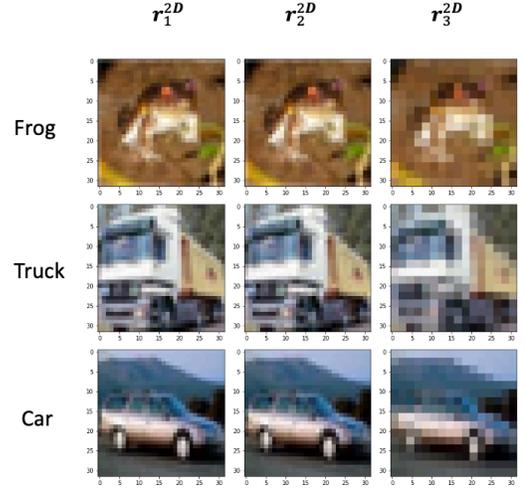

Fig. 9. Examples of a few sample representations of CIFAR-10 at three different resolution levels.

### B. Experiment Setup

For CIFAR-10 and ESC-10, we always train models with 100% of the training data. For FSDD data set, two groups of experiments are designed. The first group is the normal experiments using 100% available training data set aside to construct CNN models. In the second group, we attempt to evaluate CNN models' performance using reduced size of training data on FSDD. To do so, instead of using 100% of the training data, a much smaller portion of training data is applied in model learning. Validation data and testing data are reserved as the same as in normal experiments. For example, for FSDD, 2000 records of total 3000 records are used as training data in the first group of normal experiments, whereas in the second group (i.e., reduced training data) of experiments, only 20% of those 2000 records of data are applied as training data for constructing CNNs.

TABLE I
THE CNN MODEL ARCHITECTURE ADOPTED ON FFSD

| Layer | # Params |
|---|---|
| Conv 1D 128, 2 / Maxpooling 1D | 384 |
| Conv 1D 128, 2 / Maxpooling 1D | 32,896 |
| Conv 1D 128, 2 / Maxpooling 1D | 32,896 |
| Conv 1D 256, 2 / Maxpooling 1D | 65,792 |
| Conv 1D 256, 2 / Maxpooling 1D | 131,328 |
| Conv 1D 256, 2 / Maxpooling 1D | 131,328 |
| Conv 1D 256, 2 / Maxpooling 1D | 131,328 |
| Conv 1D 256, 2 / Maxpooling 1D | 131,328 |
| Conv 1D 256, 2 / Maxpooling 1D | 131,328 |
| Conv 1D 512, 2 / Maxpooling 1D | 262,656 |
| Conv 1D 512, 1 | 262,656 |
| Dense 128 | 917,632 |
| Dense 10 | 1,290 |
| Total # Params | 2,232,842 |

DNN architecture and setup for FSDD. With FSDD data set, sound waveforms are directly taken as input to a deep CNN model. Reported in recent work of SampleCNN [23], a CNN model that takes raw waveforms as input can achieve the classification performance to be in a par with the use of log-mel spectrograms as input to DNN models, provided that a very small filter size (such as two or three) is used in all convolutional layers. The CNN architecture given in SampleCNN is adopted in our experiments with FSDD. Table I illustrates the CNN architecture and its number of parameters for each layer. The portion of training/validation/testing data is 64%, 16% and 20% respectively for the group of normal experiments.

In our multiresolution learning for DNN models on both data sets FFSD and ESC-10, Haar wavelet transform is adopted to build multiresolution training data. Three groups of multiresolution learning DNN models are constructed on FFSD with four, five and six levels of multiresolution learning processes separately. To make sure DNN models get sufficient training and at the same time prevent overfitting, 500 total epochs and an early stopping scheme are applied. To have fair comparisons of multiresolution learning (ML) versus traditional learning (TL), for experiments with ML, the maximum epoch number for each individual resolution training phase is divided evenly from the total 500 epochs. The input size of the data is (16000, 1). Stochastic gradient descent (SGD) optimizer is adopted with a learning rate of 0.02 and momentum of 0.2 and batch size of 23. The weight initializer is set as a normal distribution with a standard deviation of 0.02 and a mean of 0. In addition, a 0.001 L2 norm weight regularizer is applied. The ReLU (Rectified Linear Unit) activation function is used for each convolutional layer and dense layer except the output layer using the softmax function. It should be noted that different slope value of ReLU function is used for different resolution level i of training data, as shown in (8). The general rule is that the finer resolution of the training data is, the smaller the slope value is [14], where the finest resolution learning phase, the slope value is 1 (i.e., the normal ReLU). The normal ReLU is also used for TL.

$$slope = 0.85 + 0.15i \qquad (8)$$

DNN architecture and setup for ESC-10. We design our new CNN model architecture for ESC-10 but adopt the same segmentation and voting scheme as in Piczak [24]. Our CNN model contains 13 layers in total as shown in Table II. The first convolutional ReLU layer consists of 80 filters of rectangular shape (3×3 size). Maxpooling layer is applied after every 2 convolutional layers with a default pool shape of 2×2 and stride of 1×1. The second to the sixth convolutional ReLU layer consist of 80 filters of rectangular shape (2×2 size). Further processing is applied through 3 dense (fully connected hidden) layers of 1000 ReLUs, 500 ReLUs and 500 ReLUs, respectively, with a softmax output layer. One dropout layer with dropout rate 0.5 are added after the third dense layer.

TABLE II
THE CNN MODEL ARCHITECTURE DESIGNED FOR ESC-10

| Layers | # Params |
|---|---|
| Conv2D 80, (3,3) | 800 |
| Conv2D 80, (2,2) | 25,680 |
| Maxpool2D | - |
| Conv2D 80, (2,2) | 25,680 |
| Conv2D 80, (2,2) | 25,680 |
| Maxpool2D | - |
| Conv2D 80, (2,2) | 25,680 |
| Conv2D 80, (2,2) | 25,680 |
| Maxpool2D | - |
| Dense 1000 | 1,201,000 |
| Dense 500 | 500,500 |
| Dense 500 | 250,500 |
| Dense 10 | 5,010 |
| Total # Params | 2,089,210 |

With the input size (60, 41, 1) for ESC-10, training is performed using SGD with a learning rate of 0.02, momentum 0.2, and batch size of 64. The same as the setup for the experiments with FSDD, 500 total epochs, and an early stopping scheme are applied. Again, for experiments with multiresolution learning, the maximum epoch number for each individual resolution training phase is obtained by dividing the total 500 epochs evenly. Also, 0.001 L2 norm weight regularizer is applied for two convolutional layers and dense layers. Different slope values of ReLU function are also applied following (8) for different resolution level i of training data. The model is evaluated in a 5-fold cross-validation regime with a single fold used as an intermittent validation set. That is, the training/validation/test data split ratio is 0.6/0.2/0.2 in the group of normal experiments. This cross-validation scheme leads to 20 different combinations of model construction and testing. Since each clip is segmented into 20 short pieces, the absolute number of training/validation/test short segments are 4800/1600/1600 respectively. Final predictions for clips are generated a probability-voting scheme.

DNN architecture and setup for CIFAR-10. We use wide Residual Networks [25] as our main network. Experiments are implemented on a popular variant WRN-28-10 which has a depth of 28 and a multiplier of 10 containing 36.5M parameters. SGD optimizer with Nesterov momentum and a global weight decay of 5x10-4. 200 training epochs are divided into 4 phases (60, 60, 40, 40 epochs respectively) to apply different learning rates (0.1, 0.02, 0.004, 0.0008) while batch size 128 is kept same. There is no validation set for training and data augmentation is not used. Instead, we evaluate top-1 error rates based on 5 runs with different seeds. While training ML2 models, $r_2^{2D}$ is the input in phase 1 and phase 2 while $r_1^{2D}$ in phase 3 and phase 4. When we train ML3 models, $r_3^{2D}$ is the input in phase 1, $r_2^{2D}$ is the input in phase 2 while $r_1^{2D}$ in phase 3 and phase 4.

## C. Results and Evaluation

With FSDD. The deep CNN models are evaluated based on CNN models constructed by the traditional single resolution learning versus the multiresolution learning using 15 different seeds for the generation of initial random weights of CNN models. To evaluate the noise robustness of trained deep CNN models on FSDD, random noise is then added to the test data. The noise set is generated using a normal probabilistic distribution with a zero mean and a scale of 0.75, with 5% noise density of the original sample length of 16000.

In the second group of experiments with small training data, we construct new CNN models by reducing the training data to only 20% of the original training data set. The evaluation results are shown in Table III, in which the accuracy for each model category is the average recognition accuracy over 15 constructed models using arbitrary seeds for models' random weight initialization.

TABLE III
AVERAGE CLASSIFICATION ACCURACY AND RELATIVE IMPROVEMENT RATIO OF ML CNNs (4, 5, 6 LEVELS) OVER TL CNNs UNDER TWO GROUPS OF EXPERIMENTS ON FSDD DATASET (100% TRAINING DATA INDICATES THE FULL TRAINING DATA USED, WHILE 20% TRAINING DATA INDICATES ONLY 20% OF THE FULL TRAINING DATA USED)

| Training data | Noise added in testing data | TL | ML | | | | | |
|---|---|---|---|---|---|---|---|---|
| | | | 4 Levels | | 5 Levels | | 6 Levels | |
| | | Acc. | Acc. | Improv. | Acc. | Improv. | Acc. | Improv. |
| 100% | N | 83.93% | 89.10% | **6.16%** | 88.19% | **5.07%** | 85.62% | **2.01%** |
| 100% | Y | 31.51% | 44.30% | **40.60%** | 44.44% | **41.04%** | 53.42% | **69.54%** |
| 20% | N | 53.82% | 66.03% | **22.69%** | 64.51% | **19.86%** | 54.03% | **0.39%** |
| 20% | Y | 25.97% | 41.21% | **58.68%** | 36.82% | **41.78%** | 39.73% | **52.98%** |

In Table. 3 and Fig. 10, as we can see, ML models always outperform TL models no matter how many total levels of multiresolution learning phases are employed in the learning process. To evaluate the constructed DNN models' performance on noisy testing data, random noise is added to test data. The performance of TL models degrades much more as we can see that the improvement ratios of ML models (with either 4-, or 5-, or 6- levels of multiresolution learning process) over TL model are all over 10%, which is significantly higher than that on clean test data. This indicates that CNN models constructed by the traditional learning are not as robust as CNN models constructed by multiresolution learning with respect to noise.

When the amount of training data is reduced in the second group of experiments, the accuracy of TL models decreases more than that of ML models, indicating that CNN models constructed by TL are more vulnerable compared to CNN models constructed by ML in small training data setting. The highest improvement over TL models is obtained by ML models in the category of 4 levels of multiresolution learning, which is 22.69%. The last row in Table III shows that, when reduced training data size and noise attack are combined, ML models demonstrate more substantial performance improvements over TL models, illustrating the benefits of multiresolution learning for CNN models in robustness enhancement with respect to noise or/and small training data.

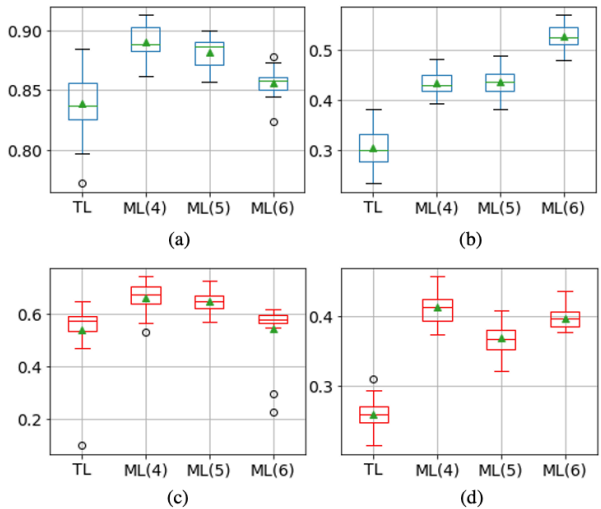

Fig. 10. Accuracy result comparison of traditional learning (single resolution) versus 4, 5, and 6 levels of multiresolution learning on FSDD. The mean accuracy for each category is given by triangle. (a)100% training data; (b)100% training data + noise; (c) 20% training data; (d) 20% training data + noise.

With ESC-10. The model is evaluated in a 5-fold cross-validation regime with a single fold used as an intermittent validation set. Since each clip is segmented into 20 short pieces, the absolute number of training/validation/test short segments is 4800/1600/1600 respectively. Final predictions for clips are generated using a probability-voting scheme. Two sets of experiments are implemented with unnormalized training data and normalized training data respectively.

We use Deepfool [8], an effective tool to systematically evaluate the adversarial robustness of CNN models on ESC-10. Deepfool is an algorithm to efficiently compute the minimum perturbations that are needed to fool DNN models on image classification tasks, and reliably quantify the robustness of these DNN classifiers using robustness metric $\rho$, where $\rho$ of a classifier M is defined as follows [9]:

$$\hat{\rho}_{adv}(M) = \frac{1}{|\mathcal{D}|} \sum_{x \in \mathcal{D}} \frac{\|\hat{p}(x)\|_2}{\|x\|_2} \qquad (9)$$

where $\hat{p}(x)$ is the estimated minimal perturbation obtained using Deepfool, and $\mathcal{D}$ denotes the test set. Basically, Deepfool is applied to the test data for each CNN model to generate adversarial perturbations. An average robustness value can be computed over the generated perturbations and original test data. The Adversarial Robustness Toolbox [25] is used to implement Deepfool.

With ESC-10 (unnormalized training data). As shown in Table IV, under normal accuracy experiments, we can see that ML models of 2 and 3 levels only slightly perform better in probability voting (PV) setting comparing to TL model. In addition, lower average test accuracy is obtained with ML models of 5 levels than TL models, which shows a degradation of 2.3%.

TABLE IV

AVERAGE CLASSIFICATION ACCURACY AND RELATIVE IMPROVEMENT RATIO OF ML MODELS (2, 3, 4, 5 LEVELS) OVER TL MODELS ON ESC-10 WITH UNNORMALIZED TRAINING DATA USING SP (SHORT SEGMENT + PROBABILITY VOTING)

| Voting scheme | TL | ML | | | | | | | |
|---|---|---|---|---|---|---|---|---|---|
| | | 2 Levels | | 3 Levels | | 4 Levels | | 5 Levels | |
| | Acc. | Acc. | Imp. | Acc | Imp. | Acc | Imp. | Acc | Imp. |
| SP | 84.06% | 84.94% | **1.05%** | 84.25% | **0.23%** | 84.06% | **0.00%** | 82.13% | **-2.30%** |

TABLE V

AVERAGE DEEPFOOL ROBUSTNESS VALUE $\rho$ OF ML (2, 3, 4, 5 LEVELS) AND TL UNDER EACH GROUP OF EXPERIMENTS ON ESC-10 WITH UNNORMALIZED TRAINING DATA

| TL | ML | | | | | | | |
|---|---|---|---|---|---|---|---|---|
| | 2 Levels | | 3 Levels | | 4 Levels | | 5 Levels | |
| $\rho$ | $\rho$ | Imp. | $\rho$ | Imp. | $\rho$ | Imp. | $\rho$ | Imp. |
| 0.277 | 0.297 | **7.22%** | 0.295 | **6.50%** | 0.303 | **9.39%** | 0.296 | **6.86%** |

TABLE VI

Average classification accuracy and relative improvement ratio of ML models (2, 3, 4, 5 levels) over TL models on ESC-10 dataset with normalized training data using SP (Short segment + Probability voting)

| Voting scheme | TL | ML | | | | | | | |
|---|---|---|---|---|---|---|---|---|---|
| | | 2 Levels | | | 3 Levels | | 4 Levels | | 5 Levels | |
| | Acc. | Acc. | Imp. | Acc | Imp. | Acc | Imp. | Acc | Imp. |
| SP | 82.00% | 81.56% | **-0.54%** | 81.88% | **-0.15%** | 83.13% | **1.38%** | 83.25% | **1.52%** |

TABLE VII

Average Deepfool Robustness value $\rho$ of ML (2, 3, 4, 5 levels) and TL under each group of experiments on ESC-10 dataset with normalized training data

| TL | ML | | | | | | | |
|---|---|---|---|---|---|---|---|---|
| | 2 Levels | | 3 Levels | | 4 Levels | | 5 Levels | |
| $\rho$ | $\rho$ | Imp. | $\rho$ | Imp. | $\rho$ | Imp. | $\rho$ | Imp. |
| 2.782 | 3.757 | **35.05%** | 4.381 | **57.48%** | 4.711 | **69.34%** | 4.993 | **79.48%** |

We compute robustness value $\rho$ for each TL and ML model. As illustrated in Table IV, when full training data are used, TL models get the lowest average $\rho$ value 0.277 compared to all different ML models, which means the perturbation needed to attack the TL models is smaller than that to the ML models. In other words, the CNN models constructed by multiresolution learning obtain better robustness against adversarial examples. Specifically, the average of $\rho$ value of the ML (4) models is about 9.4% higher than the average $\rho$ value of the TL models. The similar average $\rho$ values of the ML (2, 3, 5) models are obtained, which are about 6.5% to 7.2% higher than the average $\rho$ value of the TL models. Fig. 11 shows results from both Table IV and Table V.

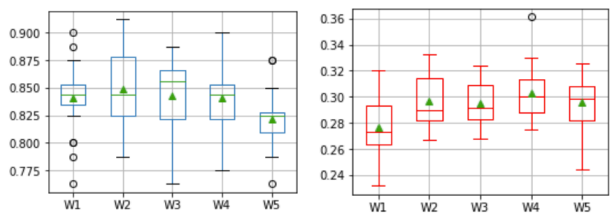

Fig. 11. Accuracy (left plot) and Deepfool robustness (right plot) result comparison of traditional learning(W1) versus 2, 3, 4 and 5 levels of multiresolution learning (W2, W3, W4, W5) on ESC-10 with unnormalized training data.

With ESC-10 (normalized training data). Experiments were also conducted with normalized input data, which fall in range [0,1] based on original spectrogram value, to explore a new robustness attack tool – AutoAttack, in addition to Deepfool. AutoAttack consists of four attacks. They are APGD-CE (auto projected gradient descent with cross-entropy loss), APGD-DLR (auto projected gradient descent with difference of logits ratio loss), FAB (Fast Adaptive Boundary Attack) and Square Attack. APGD is a white-box attack aiming at any adversarial example within an Lp-ball, FAB minimizes the norm of the perturbation necessary to achieve a misclassification, while Square Attack is a score-based black-box attack for norm bounded perturbations which uses random search and does not exploit any gradient approximation [1]. In our experiments, attack batch size 1000, epsilon 8/255 and L2 norm are applied. One special requirement of AutoAttack is softmax function in the last dense layer should be removed, which results in different sets of standard test accuracy as shown in Table VI and Table VIII (before attack). As shown in Table VI, we can see that ML (4) & ML (5) models slightly perform better comparing to TL model. In addition, lower average test accuracy is obtained with ML (2) levels than TL models, which shows a small degradation of 0.54%. Overall performance shows that normalizing data does not help improve classification accuracy.

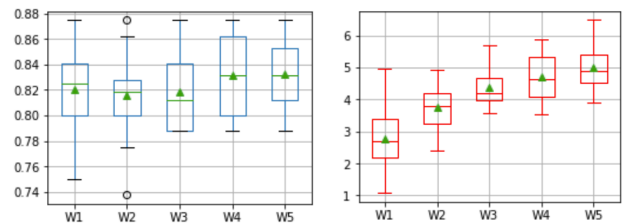

Fig. 12. Accuracy (left plot) and Deepfool robustness (right plot) result comparison of traditional learning(W1) versus 2, 3, 4 and 5 levels of multiresolution learning (W2, W3, W4, W5) on ESC-10 with normalized training data.

We further examine average Deepfool robustness value $\rho$ for each TL and ML model on normalized training data. As illustrated in Table VII, TL models get the lowest average $\rho$ value 2.782 compared to all different ML models. In general, higher robustness value $\rho$ corresponds to ML with more resolution levels. Specifically, the average of $\rho$ value of the ML (5) models is about 79.5% higher than the average $\rho$ value of the TL models. And ML (2) models achieve 35.0% higher value $\rho$ compared to TL models. In general, robustness value $\rho$ obtained with experiments on normalized training data is significantly higher than that on unnormalized training data. The results also show that the CNN models constructed by multiresolution learning demonstrate significant improvement on robustness against adversarial examples compared to TL models with normalized training data. Fig. 12 shows results from Table VI and Table VII.

| Phase (AutoAttack) | TL | ML | | | | | |
|---|---|---|---|---|---|---|---|
| | | 2 Levels | | 3 Levels | | 4 Levels | |
| | Acc. | Acc. | Imp. | Acc | Imp. | Acc | Imp. |
| Before | 74.69% | 77.19% | **3.35%** | 77.19% | **3.35%** | 77.75% | **4.10%** |
| After | 59.46% | 61.71% | **3.78%** | 61.60% | **3.60%** | 62.83% | **5.67%** |

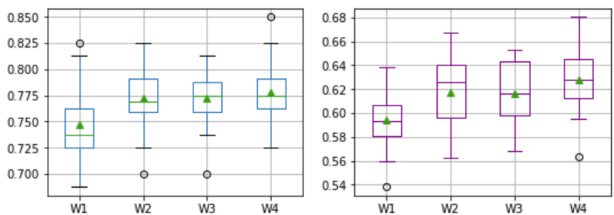

Fig. 13. Accuracy boxplot before applying AutoAttack (left plot) and Accuracy boxplot after applying AutoAttack (right plot) result comparison of traditional learning(W1) versus 2, 3, 4 and 5 levels of multiresolution learning (W2, W3, W4) on ESC-10 with normalized input data.

Table VIII and Fig. 13 illustrate AutoAttack experiment results. As we can see, removing softmax function causes some general performance degradation on standard test accuracy comparing to results in Table IV. In other words, this illustrates the effectiveness of softmax function. Different from Table IV, all ML models outperform TL models under Autoattack standard test accuracy setting (Phase -- before Autoattack). ML (4) models achieve the best improvement of 4.1% in comparison to TL models. After applying Autoattack, ML (4) models still achieve the best performance as opposed to all other models. Hence, we can draw the same overall trend as Deepfool experiments -- the ML models with more resolution levels tend to achieve better robustness performance.

With CIFAR-10. The average classification accuracy results on CIFAR-10 with WRN-28-10 models are shown in Table IX. Results indicate that TL models slightly outperform ML2 models while performance of ML3 models get a degradation of 1.31% comparing to TL models.

To evaluate the robustness of ML versus TL models, we also apply Deepfool [8] as described in section III.C. The average robustness values are reported in Table X. As we know, under the same setting, the higher robustness value indicates the better resistance to Deepfool perturbation. We find that both ML2 and ML3 models outperform TL models significantly with improvement ratio of 14.06% and 20.48% respectively. This trend can also be seen for experiments on ESC-10 dataset. The multiresolution learning models with more resolution levels, in general, achieve better Deepfool robustness.

To thoroughly investigate the robustness improvement by multiresolution learning, another black-box attack tool, one-pixel-attack [26], is further adopted in our evaluation. The one-pixel-attack tool is a method for generating one-pixel adversarial perturbations based on differential evolution (DE). We apply the untargeted attack for our models. 1000 samples out of all the successfully predicted images will be chosen randomly to test robustness. Once a perturbated sample is predicted as a wrong class, we call a success attack. The overall average attack success rate for models is obtained with 5 runs of different seeds. The results are presented in Table XI. A lower attack success rate always implies that the trained model is more robust. From Table XI, it can be seen that TL models achieve the highest attack success rate 26.0% within 1000 attacked samples. At the same time, ML2 models achieve the lowest attack success rate 24.1%, a reduction of 7.3% of attack success rate compared to TL models. ML3 models also obtain a similar result as ML2 models. To sum up, both Deepfool attack and one-pixel attack results consistently indicate that ML models are more robust than TL models under the same setting.

TABLE IX
AVERAGE CLASSIFICATION ACCURACY AND RELATIVE IMPROVEMENT RATIO OF ML MODELS (2, 3 LEVELS) OVER TL MODELS ON CIFAR-10 DATASET

| Training data | TL | ML | | | |
|---|---|---|---|---|---|
| | | 2 Levels | | 3 Levels | |
| | Acc. | Acc. | Imp. | Acc | Imp. |
| 100% | 94.86% | 94.78% | **-0.08%** | 93.62% | **-1.31%** |

TABLE X
AVERAGE DEEPFOOL ROBUSTNESS VALUE $\rho$ AND RELATIVE IMPROVEMENT RATIO OF ML MODELS (2, 3 LEVELS) OVER TL MODELS ON CIFAR-10 DATASET

| Training data | TL | ML | | | |
|---|---|---|---|---|---|
| | | 2 Levels | | 3 Levels | |
| | Robustness $\rho$ | Robustness $\rho$ | Imp. | Robustness $\rho$ | Imp. |
| 100% | 8.579 | 9.785 | **14.06%** | 10.336 | **20.48%** |

TABLE XI
AVERAGE ONE-PIXEL ATTACK SUCCESS RATE (AS RATE) AND RELATIVE IMPROVEMENT RATIO OF ML MODELS (2, 3 LEVELS) OVER TL MODELS ON CIFAR-10 DATASET

| Training data | TL | ML | | | |
|---|---|---|---|---|---|
| | | 2 Levels | | 3 Levels | |
| | AS rate | AS rate | Imp. | AS rate | Imp. |
| 100% | 26.00% | 24.10% | **7.30%** | 24.40% | **6.15%** |

TABLE XII
AVERAGE MULTI-RESOLUTION ROBUSTNESS ATTACK SUCCESS RATE (AS RATE) ON DIFFERENT COEFFICIENTS AND RELATIVE IMPROVEMENT RATIO OF ML MODELS (2, 3 LEVELS) OVER TL MODELS ON CIFAR-10 DATASET

| Coefficient | TL | ML | | | |
|---|---|---|---|---|---|
| | | 2 Levels | | 3 Levels | |
| | AS rate | AS rate | Imp. | AS rate | Imp. |
| LH | 7.80% | 6.41% | **17.83%** | 3.99% | **48.85%** |
| HL | 9.43% | 7.81% | **17.18%** | 4.34% | **53.98%** |
| HH | 3.41% | 2.48% | **27.28%** | 0.81% | **76.25%** |
| Total | 20.63% | 16.70% | **19.05%** | 9.14% | **55.70%** |

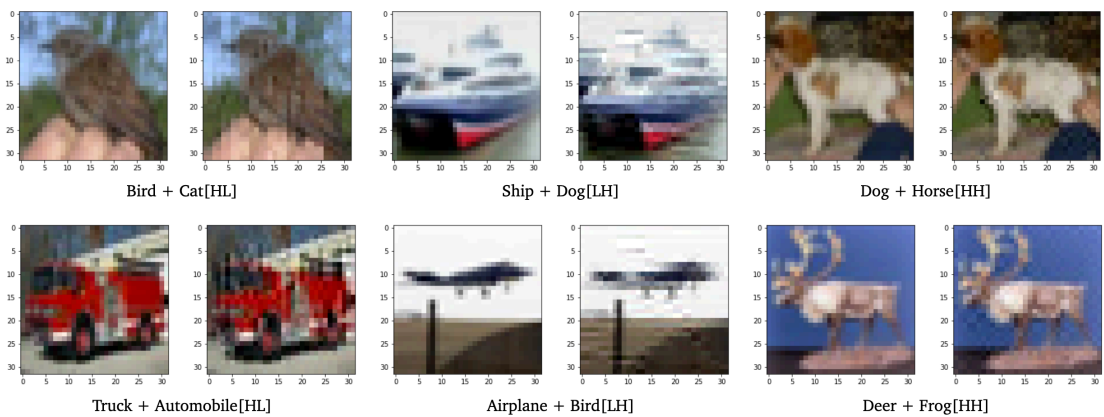

Bird + Cat[HL] Ship + Dog[LH] Dog + Horse[HH]

Truck + Automobile[HL] Airplane + Bird[LH] Deer + Frog[HH]

Fig. 14. Multi-resolution robustness attack examples. For each original image, one detailed coefficient subset among LH, HL or HH is replaced by that of another random image of different category to generate adversarial images.

Last, we propose our multi-resolution robustness attack method. To implement this attack, we apply 2D haar wavelet transformation on correct predicted test images and obtain approximation (LL), horizontal detail (LH), vertical detail (HL) and diagonal detail (HH) coefficients respectively. Then, one component among LH, HL and HH of each image is replaced randomly with the same detail component from another image of different category to generate adversarial images. The comparison between generated adversarial images and original images is shown in Fig. 13. As we can see, replacing coefficient doesn't affect when human identify different categories. Through 5 runs of attack, we obtain robustness results illustrated in Table XII. Overall, ML2 (AS rate 16.7%) and ML3 (AS rate 9.14%) both significantly outperform TL (AS rate 20.63%) while identifying perturbated images. It shows that replacing HL or LH coefficients makes it harder for models to identify perturbated images, while replacing HH coefficients is less harmful. This makes sense since HH coefficients, in general, contain detailed information of images which is less critical. In summary, TL model shows the least robustness while test images are perturbated by multi-resolution robustness attack compared to ML2 and ML3 models.

## IV. Discussions

### A. Wavelet CNNs.

Several works are reported on wavelet-based CNNs to incorporate DWT in CNNs, which can be basically classified into two categories. The first category is to use wavelet decomposition as a *preprocessing* of data to obtain a fixed level of lower resolution training data, to be employed to construct a CNN model using traditional learning process. For example, Wavelet-SRNet [27] is such a wavelet-based CNN for face super resolution. Zhao [28] focuses on experiments in ECG (1D signal) classification task using deep CNN with wavelet transform, where, again, wavelet transform is only used as a data preprocessing tool to obtain the filtered ECG signal. The second category includes schemes that embed DWT into CNN models mainly for image restoration tasks (e.g., deep convolutional framelets [29], MWCNN [30]). Our approach of multiresolution learning for deep CNN is different from those wavelet-based CNN schemes in literature in the following key aspects: First, the overall training of CNN model is a progressive process involving multiple different resolution training data from coarser versions to finer versions. In contrast, when DWT is only used for data preprocessing, the CNN training is still traditional learning, a repeated process over the single resolution training data. Second, our approach is very flexible, where users can easily select a different wavelet basis and a total levels of wavelet decomposition for a given task. In contrast, embedded DWT into CNN means that the wavelet transformation and basis used as well as the total levels of decomposition are 'hard coded' in the model, making any change difficult and error prone.

### B. Accuracy performance comparison with other schemes on ESC-10.

ESC-10 is a popular sound classification dataset which has been used by many researchers. The performance of our proposed multiresolution learning CNN is compared with other existing various schemes on ESC-10 using traditional learning, as shown in Table XIII, where our result of 5-level multiresolution learning is used.

TABLE XIII
Environment Sound Classification Performance Comparison
with Other Existing Models on ESC-10 dataset

| Year [References] | Method | Accuracy |
|---|---|---|
| 2015[24] | PiczakCNN + SP | 78.3% |
| 2015[24] | PiczakCNN + LP[1] | 80% |
| 2017[31] | SB-CNN (DA) | 79% |
| 2017[31] | SB-CNN | 73% |
| 2019[32] | CNN[2] | 73% |
| 2019[32] | TDSN[2] | 56% |
| 2019[33] | ANN[2] + Cascade | 64.5% |
| 2023[This paper] | **ML CNN + SP** | **83.25%** |

While our CNN model adopts the same segmentation and voting scheme as in the baseline work of PiczakCNN [24], our network scale is greatly reduced. PiczakCNN has about 25M trainable parameters, whereas our CNN only has about 2M parameters. That means that our CNN model has drastically saved computational resource and training time. At the same time, a better recognition result is achieved by our approach with much fewer model parameters. We note that we do not apply data augmentation (DA) in our study, while ×10 data augmentations for PiczakCNN [24] and ×5 data augmentations for SB-CNN [31] were used respectively. As we can see, our multiresolution learning CNN under short segment/probability voting setting (SP) outperforms all the other schemes/models listed in Table XIII, even though our ML CNN only used about 9.1% and 16.7% amount of training data compared to data augmentations used in PiczakCNN and SB-CNN, respectively.

### C. Accuracy performance comparison with other schemes on CIFAR-10.

In recent years, the state-of-the-art performance on CIFAR-10 has continued to improve, with new models and techniques that achieve increasingly lower error rates. In this work, we do not apply complex data augmentation method except the crop/flip on training data to focus on the effect of multiresolution learning. Table XIV presents the comparison of some classification results for this setting. As we can see, our multiresolution learning approach has achieved the higher accuracy than other schemes except for the work of [37] which improved SGD by ensembling a subset of the weights in late stages of learning to augment the standard models.

---

[1] LP: Long segment + Probability voting.

[2] Unclear whether data augmentation was used or not.

TABLE XIV
Image classification performance comparison with other existing models without applying data augmentation on CIFAR-10 dataset

| Year [References] | Network | Accuracy |
|---|---|---|
| 2021[34] | WRN-28-10 | 86.09% |
| 2020[35] | Mobilenet V2 | 92.4% |
| 2018[36] | NiN | 91.54% |
| 2021[37] | WRN-28-10 | 96.46% |
| 2021[38] | WRN-28-10 | 86.00% |
| 2023[This paper] | **WRN-28-10** | **94.78%** |

### D. Tradeoff between accuracy and robustness

Recent studies show that robustness may be at odds with accuracy in traditional learning setting (e.g., [15], [39]). Namely, getting robust models may lead to a reduction of standard accuracy. It is also stated in [15] that though adversarial training is effective, this comes with certain drawbacks like an increase in the training time and the potential need for more training data. The multiresolution deep learning, however, can overcome these drawbacks easily because it only demands the same computing epochs and the same dataset as traditional learning, as demonstrated in our experiments. Both [15] and [39] claim that adversarial vulnerability is not necessarily tied to the standard training framework but is rather a property of the dataset. This may perhaps need further study and evidence, since under robustness attack different training frameworks (e.g., traditional learning vs. multiresolution learning) could bring totally distinct results. A hypothesis is raised in [15], [39] that the biggest price of being adversarial robust is the reduction of standard accuracy. However, our results seem to suggest that multiresolution learning could significantly improve robustness with no or little tradeoff of standard accuracy.

### V. Conclusions and Future Work

In this paper, we present a new multiresolution deep learning for CNNs. To the best of our knowledge, this work represents the first study of its kind in exploring multiresolution learning in deep learning technology. We showed that multiresolution learning can significantly improve the robustness of deep CNN models for both 1D and 2D signal classification problems, even without applying data augmentation. We demonstrated this robustness improvement in terms of random noise, reduced training data, and in particular well-designed adversarial attacks using multiple systematic tools including Deepfool, AutoAttack, and one-pixel-attack. In addition, we have also proposed our systematic multi-resolution attack method for the evaluation of robustness. Our multiresolution deep learning is very general and can be readily applied to other DNN models. Our multi-resolution attack method can also be applied in general for robustness attack and adversarial training. On contrary to the recent observation in traditional single resolution learning setting, our results seem to suggest that it may not be necessary to trade standard accuracy for robustness with multiresolution learning, a definite interesting and important topic for further future research. We also plan to further investigate our approach on large-scale problems including ImageNet, and to explore various wavelet bases beyond Haar wavelet in the multiresolution learning.